\documentclass[sigconf]{aamas}

\usepackage{balance} 

\usepackage{algorithm}
\usepackage{algorithmic}
\usepackage{booktabs}
\usepackage{makecell}
\usepackage{tcolorbox}
\usepackage{xcolor}
\usepackage{amsmath}
\usepackage{array}
\usepackage{amsfonts} 
\usepackage{hyperref}

\usepackage{makecell}
\usepackage{adjustbox}

\setcopyright{none}

\title[AAMAS-2026 Formatting Instructions]{A New Perspective on Transformers in Online Reinforcement Learning for Continuous Control}

\author{Nikita Kachaev}
\authornote{Equal contribution.}
\affiliation{
  \institution{Cognitive AI Lab}
  \city{Moscow}
  \country{Russia}
}

\author{Daniil Zelezetsky}
\authornotemark[1] % 
\affiliation{
  \institution{IAI MIPT}
  \city{Moscow}
  \country{Russia}
}

\author{Egor Cherepanov}
\affiliation{
  \institution{Cognitive AI Lab, IAI MIPT}
  \city{Moscow}
  \country{Russia}}
  
\author{Alexey K. Kovalev}
\affiliation{
  \institution{Cognitive AI Lab, IAI MIPT}
  \city{Moscow}
  \country{Russia}}

\author{Aleksandr I. Panov}
\affiliation{
  \institution{Cognitive AI Lab, IAI MIPT}
  \city{Moscow}
  \country{Russia}}

\begin{abstract}
Despite their effectiveness and popularity in offline or model-based reinforcement learning (RL), transformers remain underexplored in online model-free RL due to their sensitivity to training setups and model design decisions such as how to structure the policy and value networks, share components, or handle temporal information.
In this paper, we show that transformers can be strong baselines for continuous control in online model-free RL. We investigate key design questions: how to condition inputs, share components between actor and critic, and slice sequential data for training. Our experiments reveal stable architectural and training strategies enabling competitive performance across fully and partially observable tasks, and in both vector- and image-based settings. These findings offer practical guidance for applying transformers in online RL.
\end{abstract}

\newcommand{\BibTeX}{\rm B\kern-.05em{\sc i\kern-.025em b}\kern-.08em\TeX}

\begin{document}

\pagestyle{fancy}
\fancyhead{}

\maketitle 

%%%%%%%%%%%%%%%%%%%%%%%%%%%%%%%%%%%%%%%%%%%%%%%%%%%%%%%%%%%%%%%%%%%%%%%%

\section{Introduction}

The transformer architecture~\citep{NIPS2017_3f5ee243} has become a cornerstone of modern deep learning, revolutionizing a wide array of domains such as natural language processing~\citep{devlin2019bertpretrainingdeepbidirectional}, computer vision~\citep{dosovitskiy2021imageworth16x16words}, and robotics~\citep{vima}. Its strengths include modeling long-range dependencies~\citep{trxl}, flexibly handling multimodal inputs~\citep{yang2021causalattentionvisionlanguagetasks}, and scaling effectively with model size and data~\citep{kaplan2020scalinglawsneurallanguage}.

Transformers have also gained traction in reinforcement learning (RL)~\citep{ni2023transformers,pleines2023memory,hcam,retrieval,staroverov2023fine,zelezetsky2025accelerating}, especially in offline~\citep{chen2021decision,traj_transformer, elmur, rate} and model-based RL~\citep{gendt}, where the learning problem is reframed as a sequence modeling over pre-collected datasets. These settings allow transformers to exploit their autoregressive capabilities to perform well without the challenges of online exploration. However, purely offline approaches, such as behavior cloning (BC), are constrained by limited expressivity and often fail to generalize to out-of-distribution (OOD) states~\citep{ross2011reductionimitationlearningstructured}. To mitigate this, hybrid pipelines that combine offline pretraining with online fine-tuning have been proposed~\citep{nair2021awacacceleratingonlinereinforcement}, improving adaptability while still depending on curated expert data.

\begin{figure}
    \centering \includegraphics[width=\linewidth]{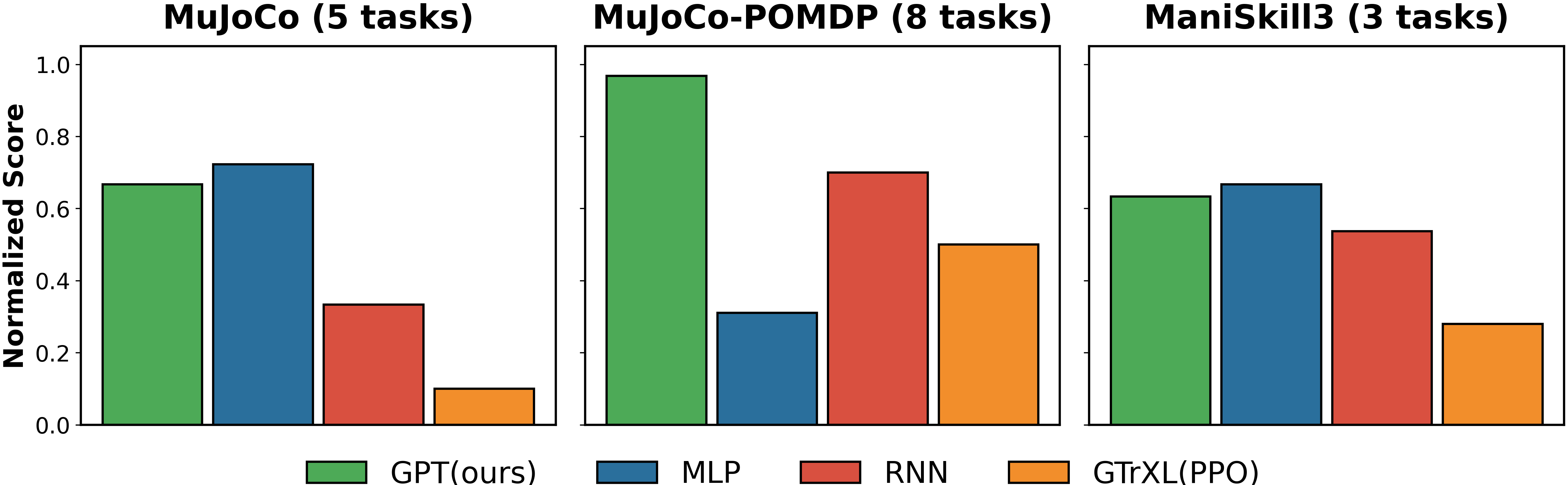}
    \caption{Transformer baselines enhanced by our findings achieve competitive performance across selected tasks. Average reward is normalized over the specified number of tasks. For MuJoCo-POMDP, we use PPO-GRU from~\citep{luo2024efficientrecurrentoffpolicyrl}.}
    \label{fig:exp_prev}
    \vspace{-5mm}
\end{figure}

Yet, both offline and hybrid RL are inherently limited by the cost, scarcity, and domain-specific nature of expert demonstrations. Online RL, in contrast, learns directly through environment interaction, which enables broader exploration and better generalization across tasks and domains. Although constrained in real-world scenarios by training time or safety concerns, online RL is especially promising in simulation-driven workflows, where large-scale data collection is feasible. These properties make it particularly well-suited for sim-to-real transfer~\citep{tobin2017domainrandomizationtransferringdeep}, which is crucial for deploying RL systems in robotics and real-world control.

Transformers are known for capturing long-term dependencies and enabling multitask generalization~\citep{traj_transformer,gtrxl,chen2021decision,gendt}, but remain underused in fully online RL. In model-free settings, adoption is limited by training instability and sensitivity to architectural and optimization design choices. Prior work~\citep{gtrxl,poliformer,relic} focuses mostly on discrete action spaces. We instead target continuous control -- a crucial yet underexplored domain, where high-dimensional and precise actions pose unique challenges.

In this work, we study how to effectively apply transformers in model-free online RL, with a focus on understanding key architectural and optimization choices. Our goal is not to propose a radically new architecture, but to surface practical insights that make transformers usable in online continuous control. We propose a unified training pipeline and systematically evaluate several critical design decisions -- such as how to condition the model, how to share parameters between actor and critic, and how to construct input sequences for temporal modeling.
We show that \textbf{transformers, though rarely used in this setting, can serve as strong baselines} for both Markov Decision Processes (MDPs) and Partially Observable MDPs (POMDPs), as well as for vector- and image-based tasks (Figure~\ref{fig:exp_prev}). Our evaluation on MuJoCo~\citep{mujoco}, its POMDP variants~\citep{pomdp_mujoco}, and ManiSkill3~\citep{maniskill3} highlights the broad applicability of our approach in continuous control. We also include the code for our experiments in the supplementary materials, enabling reproduction of the results for the proposed approaches.

\textbf{Our contributions are as follows:}
\begin{itemize}
\item \textbf{Transformer Viability in Online RL:} We demonstrate that with the right training setup, transformers are competitive across MDP and POMDP locomotion tasks and both vector- and image-based robotic control.
\item \textbf{Insights and Recommendations:} We distill actionable guidelines from extensive experiments to support stable and efficient transformer training in online RL.
\end{itemize}

\section{Transformers in RL}
\paragraph{\textbf{Offline RL}}
\citet{chen2021decision} reformulates RL as sequence modeling problem, introducing the Decision Transformer (DT), which generates actions autoregressively from past states, actions, and returns-to-go. This approach replaces value estimation with supervised learning on offline data, achieving strong performance in offline RL.
\citet{traj_transformer} extends this idea by introducing the Trajectory Transformer (TT), which incorporates learned dynamics for planning and bridges imitation and model-based RL.
\citet{mgdt} further adapts DT to train a single agent across multiple environments.
Despite strong offline results, these methods face key limitations -- most notably, their reliance on expert data, which is costly, hard to collect, or unsafe in real-world settings.
Even with expert data, offline RL models often struggle to generalize due to limited exploration and shifts in the environment distribution.

\paragraph{\textbf{Hybrid Methods}}
To mitigate the reliance on expert demonstrations and improve generalization beyond static datasets, training must incorporate online interaction. Online Decision Transformer (ODT)~\citep{online_dt} moves in this direction by combining offline likelihood maximization with online updates.
SMART~\citep{sun2023smart} successfully applies transformers in online RL by using self-supervised offline pretraining, which enables more stable online learning.
Hybrid methods aid exploration and adaptation but still rely on offline data, limiting their general applicability.

\paragraph{\textbf{Online RL}}
Fully online training removes the need for expert data but remains less explored and more challenging for transformers than offline and hybrid approaches.
\citet{dtqn} trains a transformer from scratch in the online setting using a Deep Q-Network (DQN)-like algorithm, achieving better results than Deep Recurrent Q-Network (DRQN)~\citep{drqn} on POMDP tasks.
While based on DQN, Deep Transformer Q-Network (DTQN) is limited to discrete action spaces, and thus cannot directly handle continuous control.
\citet{gtrxl} introduced Gated Transformer-XL (GTrXL), a RL-focused variant of Transformer-XL~\citep{trxl}, adding gating and identity map reordering. GTrXL proved effective in memory-intensive tasks~\citep{pleines2023memory}.
The gating mechanism stabilize large multi-layered transformers by allowing them to bypass the attention and feed-forward components within each block. This enables the model to dynamically control how much information is transformed versus passed through unchanged, effectively regulating the flow of content around attention processing.
Another effort to apply transformers in online RL involves the Recurrent Linear Transformer (ReLiT) and its gated variant, AGaLiTe~\citep{relit}, which reduce quadratic complexity via context-independent inference, but may underperform on tasks requiring richer temporal modeling.

\begin{figure}[t]
\centering
\includegraphics[width=\linewidth]{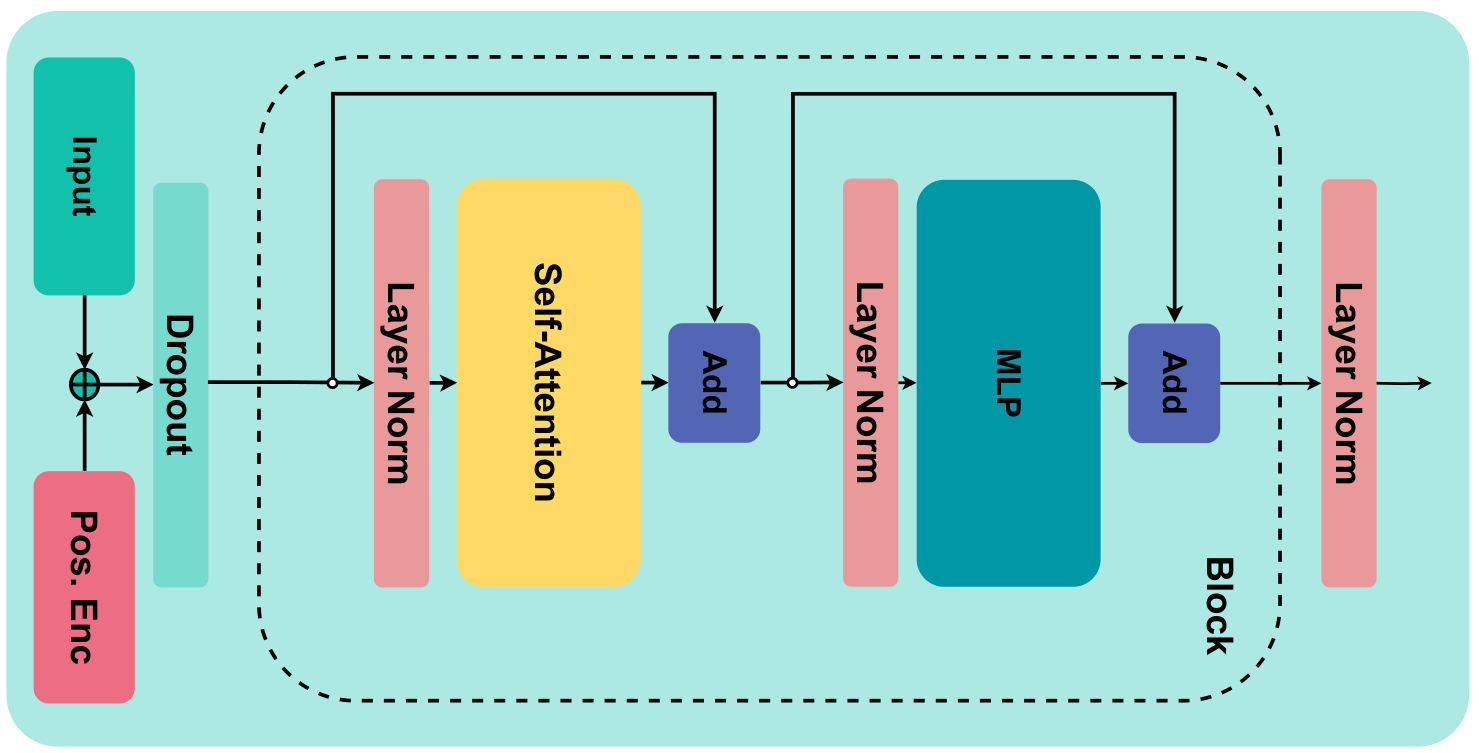}
\caption{Scheme of the GPT-2-like block used as a backbone during all the experiments in this work.}
\label{fig:gpt_scheme}
\vspace{-15pt}
\end{figure}

While these works offer promising directions, transformer-based online RL remains underexplored. In this paper, we take a systematic approach to investigating how transformers can be made competitive in fully online model-free RL, identifying key training and architectural principles for their effective use.

%%%%%%%%%%%%%%%%%%%%%%%%%%%%%%%%%%%%%%%%%%%%%%%%%%%%%%%%%%%%%%%%%%%%%%%%

%%%%%%%%%%%%%%%%%%%%%%%%%%%%%%%%%%%%%%%%%%%%%%%%%%%%%%%%%%%%%%%%%%%%%%%%

\section{Methodology}
To show that transformers can be a strong baseline for continuous control tasks, we conduct a series of experiments guided by key research questions, each addressing a core design challenge. For each, we distill practical insights, then combine the best strategies into a unified setup to demonstrate competitive performance.

\textbf{The research questions (RQs) we investigate are:}
\begin{enumerate}
    \item How does input conditioning affect transformer performance?
    \item What is the impact of sharing the backbone between actor and critic?
    \item How does data slicing influence training?
\end{enumerate}

We use MLPs as a standard baseline for MDP tasks and CNNs for image-based settings. RNNs offer a lightweight sequential alternative to transformers, making them a useful point of comparison to assess whether the added complexity of transformers is warranted.

\paragraph{\textbf{Environments}}
We validate our best transformer configuration by comparing it to MLP and RNN baselines in both MDP and POMDP settings, and demonstrate that it generalizes from vector-based to more challenging image-based tasks. Our evaluation spans three environment suites: MuJoCo~\citep{mujoco} for standard continuous control tasks, using environments such as \texttt{HalfCheetah}, \texttt{Ant}, \texttt{Hopper}, \texttt{Humanoid}, \texttt{Walker}, \texttt{Pusher}, and \texttt{Reacher}; ManiSkill3~\citep{maniskill3} for robotic manipulation, with vector-based tasks like \texttt{PushT}, \texttt{Pick Cube}, \texttt{TriFingerRotateCube}, as well as image-based tasks such as \texttt{PushCube}, \texttt{PickCube}, and \texttt{PokeCube}; and MuJoCo-POMDP~\citep{pomdp_mujoco}, which introduces partial observability by masking velocity or position to highlight the importance of temporal modeling in sequential decision-making. This setup enables a comprehensive assessment of transformer performance across diverse observation modalities and control challenges.

\begin{figure*}[h]
    \centering
    \includegraphics[width=1\linewidth]{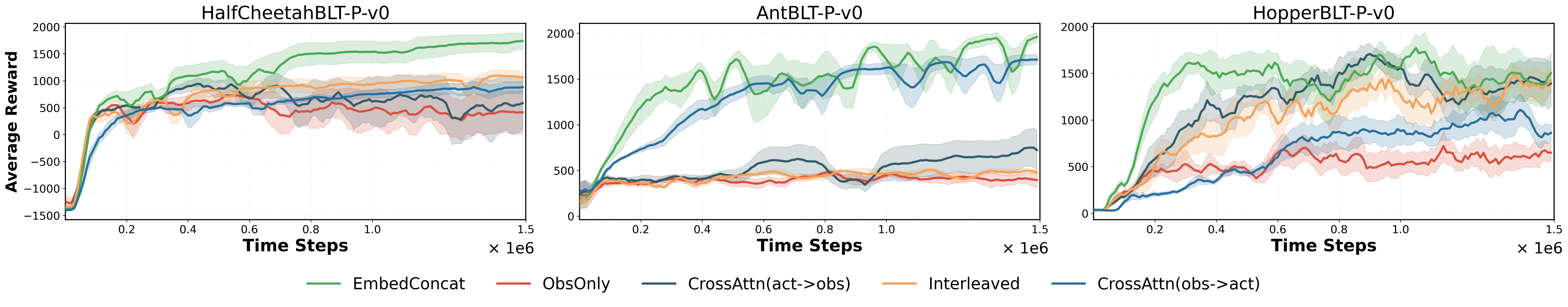}
    \vspace{-20pt}
    \caption{Performance of transformer conditioning methods on MuJoCo-POMDP tasks, where velocity is masked to induce partial observability. The two \texttt{\textbf{CrossAttn}} variants differ in sequence input order.}
    \label{fig:exp_conditioning}
\end{figure*}

\paragraph{\textbf{Transformer Baselines}}
To train transformer-based models, we use both on-policy and off-policy algorithms: Proximal Policy Optimization (PPO)~\citep{schulman2017proximalpolicyoptimizationalgorithms}, Twin Delayed Deep Deterministic Policy Gradient (TD3)~\citep{td3}, and Soft Actor-Critic (SAC)~\citep{sac}. Our goal is to identify training patterns that that remain consistent across algorithms. For ManiSkill3 experiments, we use the official SAC and PPO implementations. For MuJoCo MDP and POMDP tasks, we use CleanRL PPO and TD3 implementations~\citep{huang2022cleanrl} using default parameters.
We use a transformer decoder (Figure~\ref{fig:gpt_scheme}) inspired by GPT-2~\citep{radford2019language}, with key modifications: pre-layer normalization from~\citep{gtrxl}, GELU activation instead of ReLU in the feed-forward layers, and integration with FlashAttention~\citep{dao2023flashattention2fasterattentionbetter} for efficient training. This backbone is used across all experiments and referred to as ``GPT'', with TD3-GPT, SAC-GPT, and PPO-GPT denoting the training algorithm used. 
Full model and training configurations are listed in Appendix A in Tables~\ref{opt_params}, \ref{tab:params_sac}, and \ref{tab:params_all}.

\paragraph{\textbf{Experimental protocol}}
For each experiment, we conducted three runs per agent with different random initializations and performed evaluation during training using 100 random seeds. The results are presented as the mean episodic reward or success rate $\pm$ the standard error of the mean.

%%%%%%%%%%%%%%%%%%%%%%%%%%%%%%%%%%%%%%%%%%%%%%%%%%%%%%%%%%%%%%%%%%%%%%%%

\section{Experiments and Results}
\subsection{RQ1: How does transformer conditioning affect performance?}
\label{sec:conditioning}

A key challenge in applying transformers to online RL is determining how to condition the model on relevant inputs. In this section, we investigate how different conditioning strategies affect performance in partially observable settings. Specifically, we use the MuJoCo-POMDP benchmark~\citep{pomdp_mujoco}, where velocity information is masked to induce partial observability and emphasize the importance of sequential processing.

We evaluate four common conditioning methods with TD3-GPT, where the transformer receives a context window and predicts the action $a_t$ from the final hidden state 
$h_t$. These strategies (Table~\ref{tab:code_style_equations}) reflect standard choices in transformer-based RL models and vary in the type and structure of information given to the model:

\begin{itemize}
    \item \texttt{\textbf{ObsOnly}} feeds a sequence of past observations into the transformer; $a_t$ is predicted from the final embedding.
    \item \texttt{\textbf{Interleaved}} extends this by including previous actions in the sequence, keeping prediction from the last observation.
    \item \texttt{\textbf{EmbedConcat}} encodes observations, actions, and rewards separately, then concatenates them into a single vector before feeding the sequence into the transformer. The action is predicted from the last combined token.
    \item \texttt{\textbf{CrossAttn}} uses a two-layer transformer: the first applies self-attention to actions, the second cross-attends to observations. The final token is used for $a_t$ prediction.
\end{itemize}

\begin{figure*}[t!]%[t]
    \centering
    \includegraphics[width=1.0\linewidth]{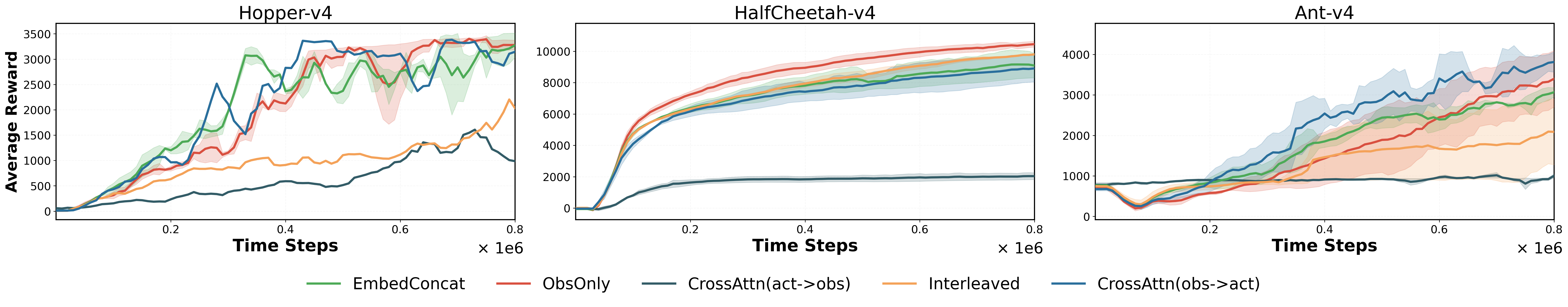}
    \vspace{-20pt}
    \caption{Conditioning experiments for MDP version of MuJoCo tasks. Using observations only is enough for tasks with Markovian properties.}
    \label{fig:rq1_mdp}
\end{figure*}

These methods represent growing contextual richness and complexity, allowing us to study how various forms of temporal and multimodal information affect performance in POMDPs.

\begin{figure}[b!]
    \centering
    \includegraphics[width=\linewidth]{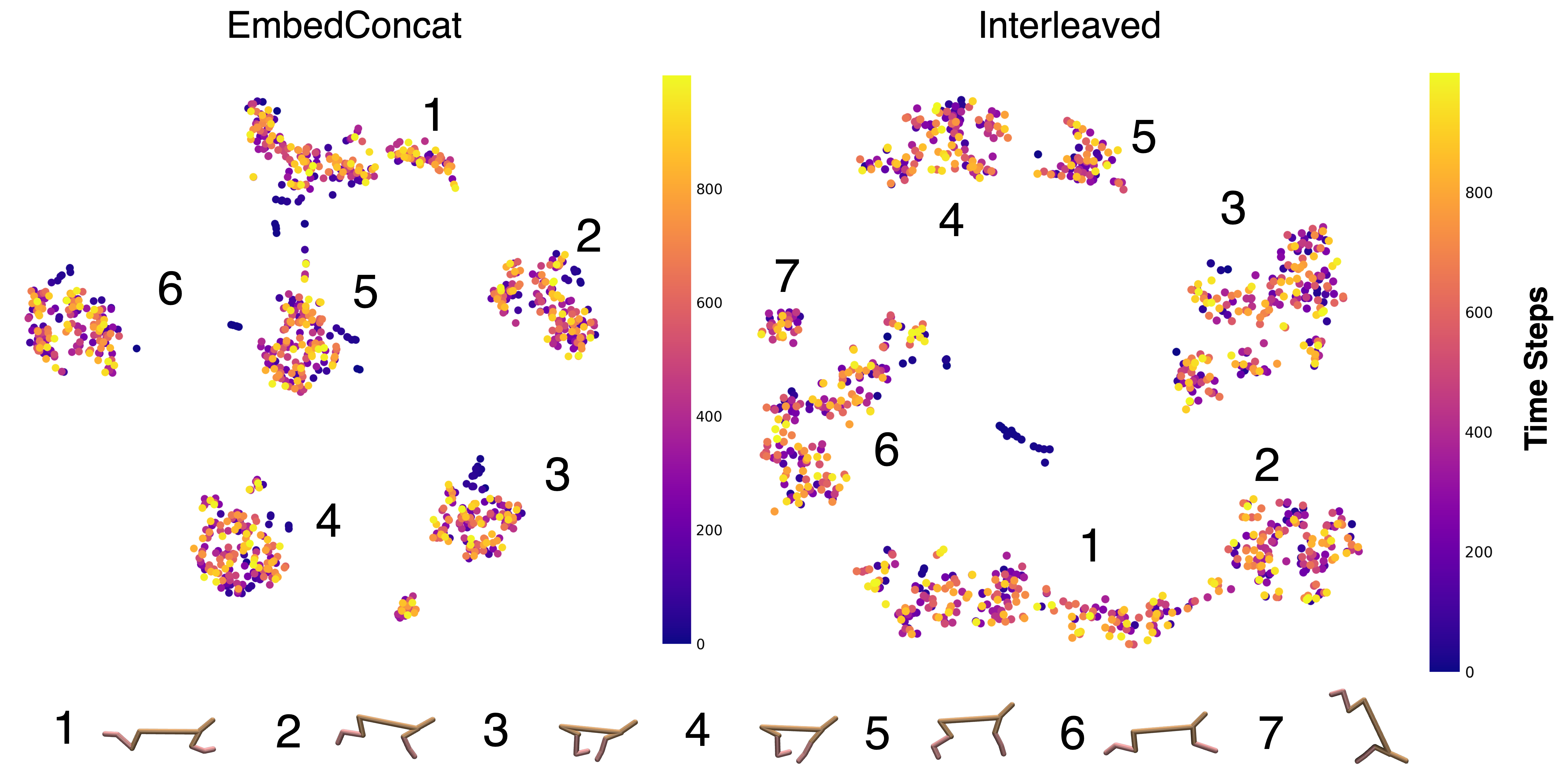}
    \caption{t-SNE visualization of action representations from the transformer on \texttt{HalfCheetah}. Clusters with \texttt{\textbf{EmbedConcat}} appear more structured.}
    \label{fig:latent_space}
\end{figure}

\paragraph{\textbf{Interpretation}}
Figure~\ref{fig:exp_conditioning} highlights two key findings. First, adding more types of information (actions and rewards) consistently improves performance in partially observable settings. Second, the \texttt{\textbf{EmbedConcat}} method proves most effective -- not only for incorporating multimodal inputs, but also for its training stability in online RL. Unlike \texttt{\textbf{CrossAttn}} or \texttt{\textbf{Interleaved}}, which rely on complex attention or position-specific queries, \texttt{\textbf{EmbedConcat}} simplifies processing by flattening and fusing all modalities upfront, making optimization easier under online training instability. Its strong performance stems from two factors: (1) it includes rewards and actions -- critical in POMDPs alongside observations, and (2) it merges each triplet (observation, action, reward) into a single embedding, yielding a homogeneous sequence. This allows the attention mechanism to focus purely on temporal dependencies. In contrast, \texttt{\textbf{Interleaved}} mixes heterogeneous inputs, complicating both temporal alignment and modality separation, which may hinder learning stability.

\begin{table}[b]
\vspace{-10pt}
\centering
\begin{tabular}{l|l}
\toprule
\textbf{Method} & \textbf{Equations} \\
\midrule

\textbf{\texttt{ObsOnly}} &
\makecell[tl]{
$ \mathtt{seq = [o_{t-M+1}, \dots, o_t]} $\\[-2pt]
$ \mathtt{h_t = SelfAttention(seq)[-1]} $
} \\
\midrule

\textbf{\texttt{Interleaved}} &
\makecell[tl]{
$ \mathtt{seq = [o_{t-M+1}, a_{t-M+1}, \dots, a_{t-1}, o_t]} $\\[-2pt]
$ \mathtt{h_t = SelfAttention(seq)[-1]} $
} \\
\midrule

\textbf{\texttt{EmbedConcat}} &
\makecell[tl]{
$ \mathtt{e_n = concat(emb(o_n), emb(a_{n-1}), emb(r_n))} $\\[-2pt]
$ \mathtt{seq = [e_{t-M+1}, \dots, e_t]} $\\[-2pt]
$ \mathtt{h_t = SelfAttention(seq)[-1]} $
} \\
\midrule

\textbf{\texttt{CrossAttn}} &
\makecell[tl]{
$ \mathtt{seq_x = [a_{t-M+1}, \dots, a_{t-1}]} $\\[-2pt]
$ \mathtt{seq_y = [o_{t-M+1}, \dots, o_t]} $\\[-2pt]
$ \mathtt{z_t = SelfAttention(seq_x)} $\\[-2pt]
$ \mathtt{h_t = CrossAttention(z_t,\, seq_y)[-1]} $
} \\
\bottomrule
\end{tabular}

\caption{Transformer conditioning strategies. We explored these strategies on both MDP and POMDP versions of MuJoCo environments.}
\label{tab:code_style_equations}
\end{table}

Similar experiments with the MDP versions of the MuJoCo environments (Figure~\ref{fig:rq1_mdp}) show that for tasks with Markovian properties, there is no need to use \texttt{\textbf{EmbedConcat}}, and \texttt{\textbf{ObsOnly}} is sufficient for successful training, but similar to the POMDP versions, the \texttt{\textbf{EmbedConcat}} method also shows strong performance on MDP.

To better understand model behavior, we visualized the transformer's final hidden states $h_t$ using t-SNE, along with the corresponding predicted actions (Figure~\ref{fig:latent_space}).
Well-performing models showed clearly separated clusters in latent space, each linked to distinct learned behaviors. The \texttt{\textbf{EmbedConcat}} method, which achieved the best results (Figure~\ref{fig:exp_conditioning}), also exhibited the most structured and interpretable clustering. We present visualizations for \texttt{\textbf{EmbedConcat}} and \texttt{\textbf{Interleaved}} as representative examples.

\begin{tcolorbox}[colback=white!3, colframe=white!60!black, boxrule=0.7pt, arc=4pt, left=4pt, right=4pt, top=4pt, bottom=4pt]
\textbf{Practical Takeaway:} For POMDPs, transformer performance improves when embeddings of observations, actions, and rewards are combined into a single input vector. In MDPs, observations alone are sufficient.
\end{tcolorbox}

%%%%%%%%%%%%%%%%%%%%%%%%%%%%%%%%%%%%%%%%%%%%%%%%%%%%%%%%%%%%%%%%%%%%%%%%
\subsection{RQ2: How does actor–critic backbone sharing affect training stability and efficiency?}
\label{sec:share_separate_transformer}
\begin{figure}[b!]
    \centering
    \includegraphics[width=1.0\linewidth]{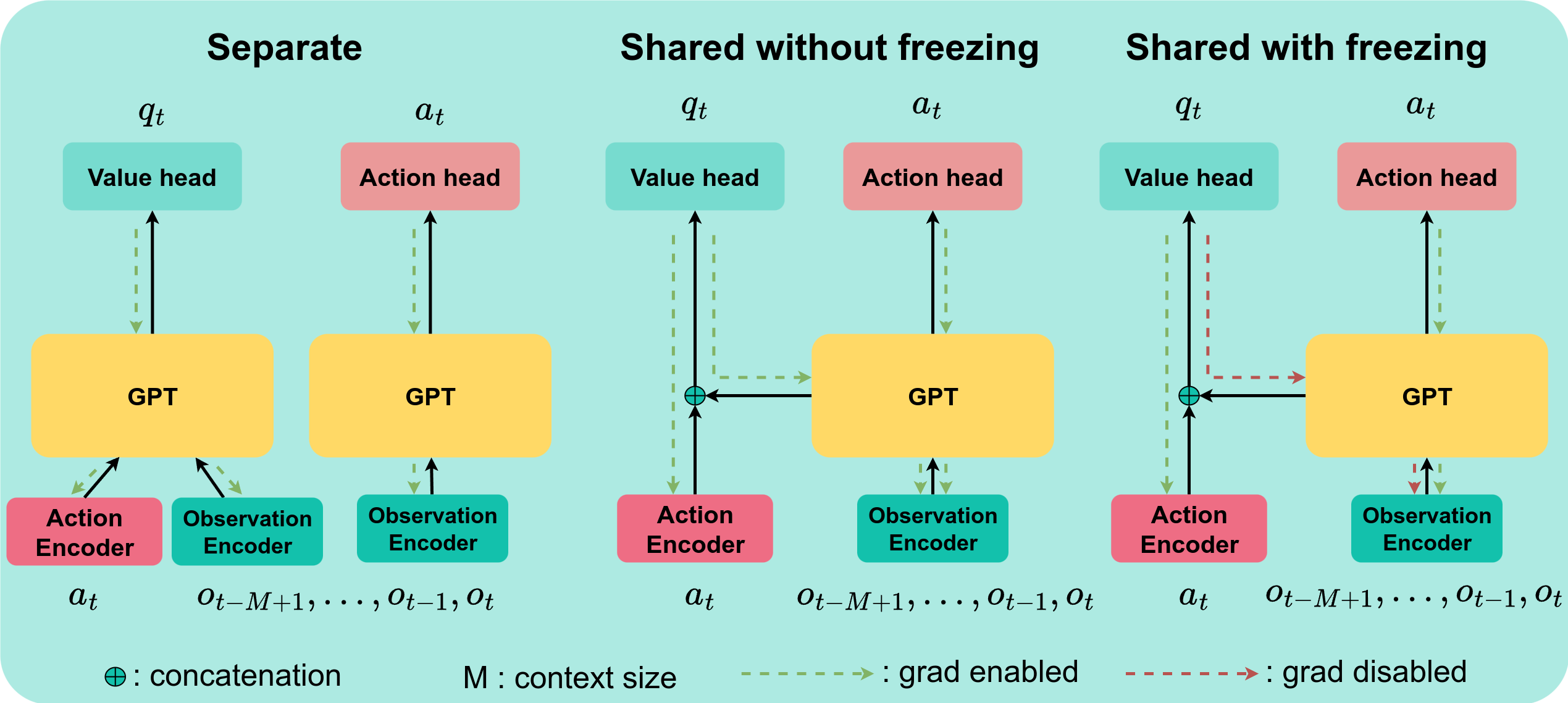}
    \caption{Visual illustration of transformer-based agent with separated and shared architectures. In the frozen variant, gradients from the critic are not propagated through the backbone and observation encoder (red dashed line).}
    \label{fig:transformer_architecture}
\end{figure}

We use TD3-GPT, where a transformer encodes observation sequences as in the \texttt{\textbf{ObsOnly}} setup. To reduce parameters without sacrificing performance, we investigate whether the actor and the critic can share the transformer backbone.
To isolate architectural effects, we use non-terminating MDPs where observations are sufficient.
Figure~\ref{fig:transformer_architecture} shows the tested sharing strategies:
\begin{enumerate}
    \item \textbf{Separate:} the actor and the critic use independent transformer backbones.
    \item \textbf{Shared without freezing:} shared transformer is updated only by the actor; the critic gradients are blocked.
    \item \textbf{Shared with freezing:} the actor and the critic share the same transformer, updated by both gradients.
\end{enumerate}

\begin{figure*}[t]
    \centering
    \includegraphics[width=1\linewidth]{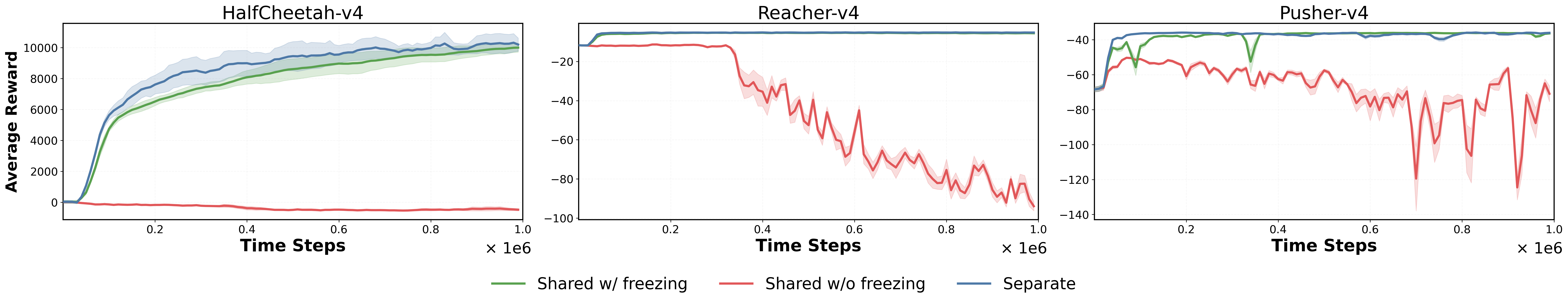}
    % \vspace{-20pt}
    \caption{Comparison of TD3-GPT backbone sharing methods. ``Separate'' and ``shared with freezing'' setups perform similarly, while removing freezing causes collapse due to gradient conflict.}
    \label{fig:exp_3}
    % \vspace{-10pt}
\end{figure*}

\begin{figure}[b]
    \centering
    \includegraphics[width=\linewidth]{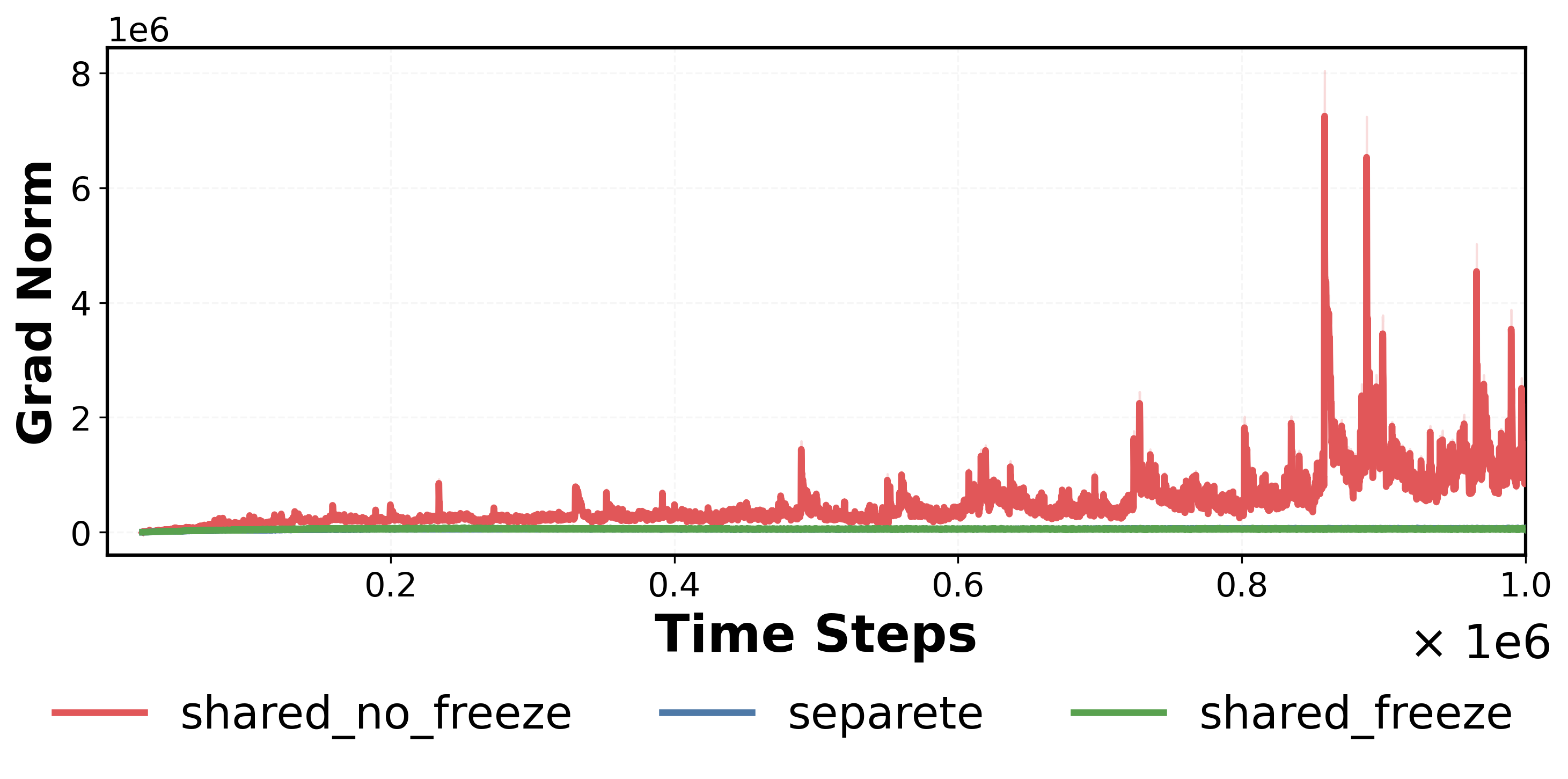}
    \vspace{-10pt}
    \caption{Gradient norms for different backbone sharing methods on \texttt{HalfCheetah}. Freezing the shared backbone during critic updates stabilizes training; removing it leads to gradient explosions. Blue curve lies beneath the green.}
    \label{fig:grad_norm}
\end{figure}

\textbf{Interpretation.}
Figure~\ref{fig:exp_3} shows that using separate backbones for the actor and the critic ensures stable training, while sharing a backbone degrades performance unless the transformer is frozen during critic updates. This indicates conflicting gradient signals between actor and critic: the actor maximizes rewards, while the critic minimizes Temporal Difference error -- objectives that may conflict. These opposing gradients can destabilize learning or suppress updates due to gradient explosion.
To test this, we logged and plotted gradient norms (Figure~\ref{fig:grad_norm}). In the ``shared without freezing'' setup, gradient norms grow continuously to extreme values. In contrast, the other two setups maintain stable updates by avoiding actor–critic interference. Similar issues of training instability between the actor and the critic are discussed by~\citet{iclr_share} and~\citet{ppg}.

\begin{tcolorbox}[colback=white!3, colframe=white!60!black, boxrule=0.7pt, arc=4pt, left=4pt, right=4pt, top=4pt, bottom=4pt]
\textbf{Practical Takeaway:} Sharing a transformer between actor and critic in off-policy RL causes gradient interference and instability. Using separate backbones improves stability but increases cost. Freezing the shared transformer during critic updates provides a stable and efficient compromise.
\end{tcolorbox}

%%%%%%%%%%%%%%%%%%%%%%%%%%%%%%%%%%%%%%%%%%%%%%%%%%%%%%%%%%%%%%%%%%%%%%%%

\subsection{RQ3: How does input data slicing affect the training of transformers?}
\label{sec:batch_making}
Due to the transformer's sequential nature, its use in online RL requires careful handling of its output sequence. Broadly, there are two main ways to process it:

\textbf{Method 1:} predict the action or Q-value only from the last hidden state processed by the transformer. 
\textbf{Method 2:} predict actions or Q-values from every hidden state in the sequence.

\begin{figure*}[t!]
    \centering
    \includegraphics[width=1\linewidth]{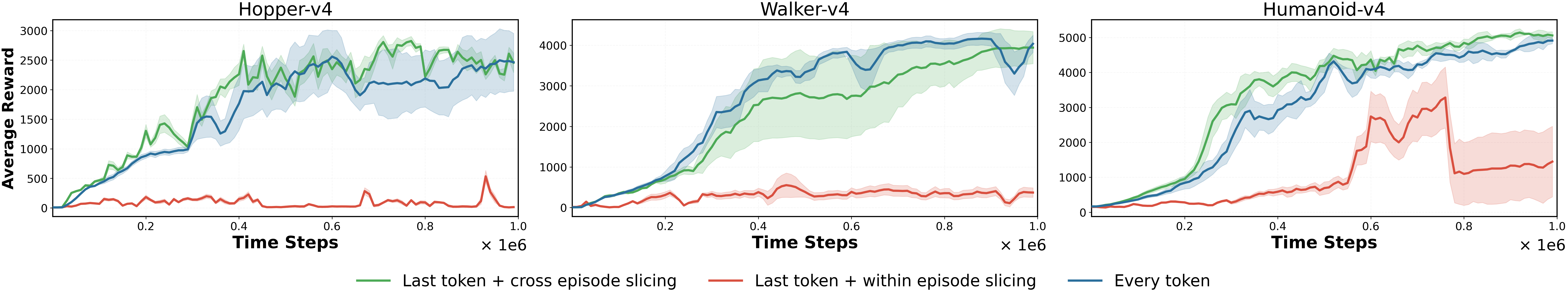}
    \caption{Evaluation performance of three ways of slicing data. Last token methods use last token to predict action during training, while Every token method trains transformer to make action prediction from every token.}
    \label{fig:data_proc}
\end{figure*}

The transformer output \texttt{(batch, context, hidden\_dim)} in \textbf{Method 1} is sliced via \texttt{[:,-1,:]} to extract the current observation's embedding. The actor and the critic thus learn from the full input sequence but predict only from the last token, requiring that the final position contains the full episode state. \textbf{Method 2} trains on all hidden states, potentially yielding better gradients and faster learning.
%, but at higher computational cost, as gradients are computed for each element.

Method 1 is more sensitive to how replay buffer data is sliced, since predictions rely solely on the current state's embedding -- the last input token. With fixed context length during training, the model implicitly assumes all sequences match this size. Otherwise, it may act suboptimally due to out-of-context evaluation, performing well only once the episode reaches the training context length.

\begin{figure}[b]
    \centering
    \includegraphics[width=1\linewidth]{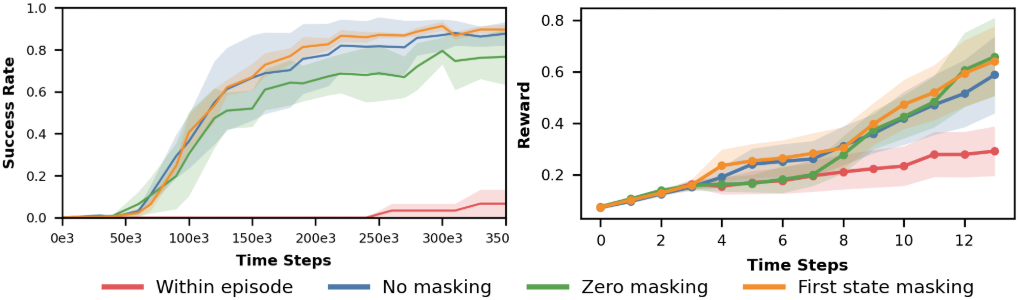}
     \caption{The return comparison of the slicing strategies on the first 14 steps in the \texttt{PickCube} environment (right) and their associated average success rate (left). }
     \label{fig:2in1}
\end{figure}

An effective data handling strategy should ideally preserve the advantages of both methods while mitigating their limitations. We argue that Method~1, when combined with cross-episode slicing, offers such a compromise: it supports stable and efficient training with convergence behavior comparable to Method~2. To better understand this process, consider Figure~\ref{fig:batch_making_scheme}. Using Method~1, we can slice state sequences either within or across episodes. At the start of an episode, the agent interacts with the environment and accumulates observations until the context length $C$ is reached (e.g., $C{=}3$ in the figure).
As shown in the left part of Figure~\ref{fig:batch_making_scheme}, the within-episode approach collects data from the episode start but skips training on the first $C-1$ steps. This gap can cause poor early-episode behavior, which is critical in tasks such as \texttt{Humanoid}, \texttt{Walker}, \texttt{Hopper}, and even ManiSkill3 (see Figure~\ref{fig:2in1}). The issue is that early observations (e.g., $o_1$, $o_2$) never appear at the final token position, leaving the agent untrained to act at those steps. The longer the context, the more pronounced the problem. Cross-episode slicing (Figure~\ref{fig:batch_making_scheme}, right) solves this by allowing input sequences to span across episode boundaries. The model still predicts from the last token, but now includes early-episode data in context, enabling better learning from the first $C{-}1$ steps.

We evaluate this in three training strategies:
\begin{itemize}
    \item \textbf{``Every token''}: make predictions at every token without applying cross-episode slicing, as the agent is trained to operate over all timesteps.
    \item \textbf{``Cross-episode''}: make predictions only at the last token while applying cross-episode slicing to ensure balanced learning across all states within an episode.
    \item \textbf{``Within-episode''}: make predictions only at the last token, while starting data collection only after the required context length has been reached.
\end{itemize}

\begin{figure}[b]
    \centering
    \includegraphics[width=\linewidth]{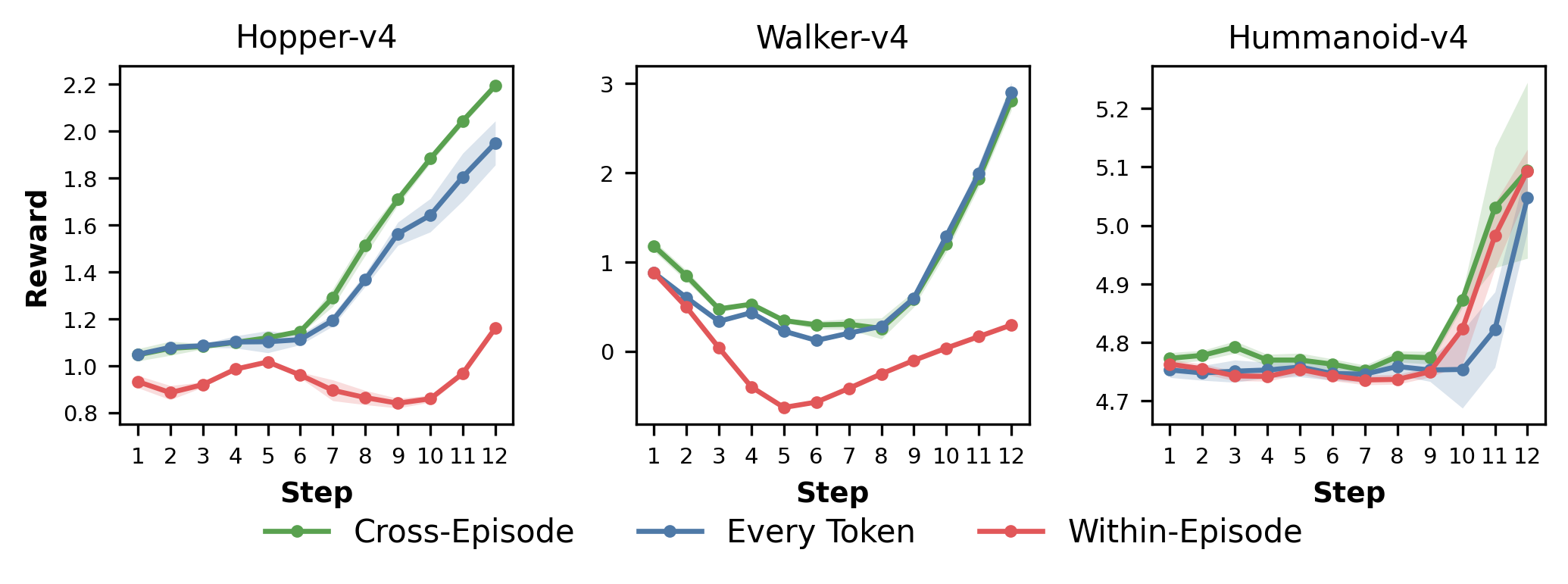}
    \caption{Reward received during first 12 steps with different ways of training. ``Cross-episode'' slicing reaches similar / better results compared with ``Every token'' approach.}
    \label{fig:all3_horizontal}
\end{figure}

\textbf{Interpretation.} Figure~\ref{fig:data_proc} shows that both ``Every token'' and ``Cross-episode'' enable effective learning, while ``Within-episode'' impairs performance despite Method~1's efficiency. To test if ``Cross-episode'' slicing improves early behavior, we measure rewards over the first 12 steps with context length 10. As shown in Figure~\ref{fig:all3_horizontal}, ``Within-episode'' underperforms early, while ``Cross-episode'' resolves this without the overhead of training on every token.

The application of the cross-episode slicing approach demonstrated promising results, stabilizing and improving training without the need to train the agent through predictions at every token in the transformer output sequence. Nevertheless, this question warrants further investigation, particularly through exploring alternative masking strategies. We considered three strategies of how to work with cross-episode slices:
\begin{figure}[h]
    \centering
    \includegraphics[width=\linewidth]{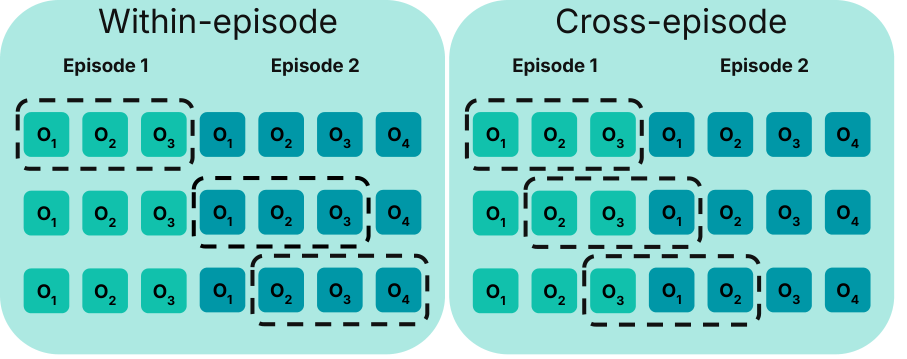}
    \caption{Within-episode slicing (left) vs. cross-episode slicing (right). The former stores data only after reaching the full context, while the latter slices sequences across episodes.}
    \label{fig:batch_making_scheme}
\end{figure}

\begin{enumerate}
    \item Ignore masking and let the model learn the boundaries that separate the end of the previous episode and the beginning of the new one.
    \item Mask irrelevant observations with zero vector.
    \item Mask irrelevant observations with the first available observation from the current episode.
\end{enumerate}

\begin{figure}[b]
    \centering
    \includegraphics[width=1\linewidth]{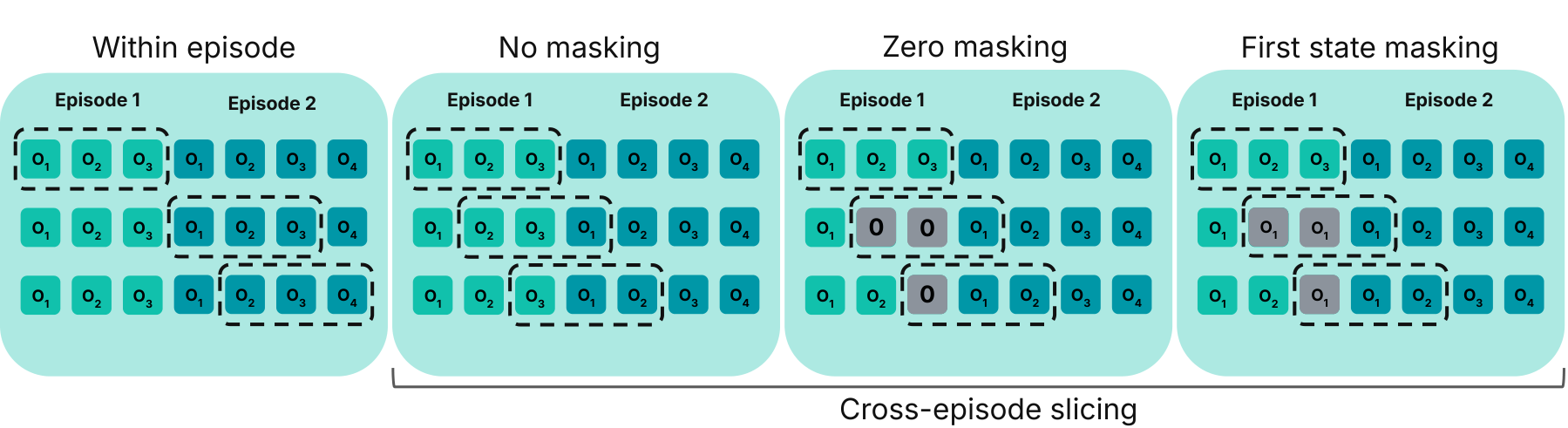}
    \caption{Four approaches of data slicing. Cross-episode slicing is represented by three ways of masking.}
    \label{22}
\end{figure}

In the additional experiment, we have tested four approaches of data-slicing, depicted on the Figure~\ref{22}: 
1) slicing within the episode (left); 2) cross-episode slicing without additional masking (the second from the left); 3) cross-episode slicing with zero mask (the third from the left); 4) cross-episode slicing with the first available observation mask (the right one).

Figure~\ref{fig:2in1} (left) supports the findings presented in the main text. Cross-episode slicing enables the transformer to successfully learn even in the ManiSkill3 \texttt{PickCube} environment. Among the three cross-episode slicing variants, the most effective are zero padding and the first available observation padding. With these results in hand, we can analyze the agent’s behavior over the first 14 steps in the environment (given a training context length of 10).

Figure~\ref{fig:2in1} (right) illustrates the agent's average reward over the first 14 steps of an episode, averaged across 100 seeds in \texttt{PickCube}. The figure supports the conclusion that the cross-episode slicing improves action quality during the initial 9 steps of the environment, compared to within-episode slicing, which lacks this mechanism. Notably, both the first observation masking and cross-episode masking approaches enhance the agent’s reward at the beginning of the episode, thereby contributing to improved overall episode performance.

Zero-padding, in turn, yields the weakest performance among the three cross-episode slicing methods (which is consistent with the results shown in Figure~\ref{fig:2in1} (left)). However, it still enables reward improvement in later steps. In contrast, the within-episode slicing approach fails to meet the challenge, and the agent continues to perform inefficiently even when sufficient context becomes available.

\begin{tcolorbox}[colback=white!3, colframe=white!60!black, boxrule=0.7pt, arc=4pt, left=4pt, right=4pt, top=4pt, bottom=4pt]
\textbf{Practical Takeaway:} When predicting from the last token, cross-episode slicing is essential for early-episode learning. Predicting from every token avoids this but adds cost without improving performance or stability.
\end{tcolorbox}

%%%%%%%%%%%%%%%%%%%%%%%%%%%%%%%%%%%%%%%%%%%%%%%%%%%%%%%%%%%%%%%%%%%%%%%%
\begin{figure*}[t!]
    \centering
    \includegraphics[width=1\linewidth]{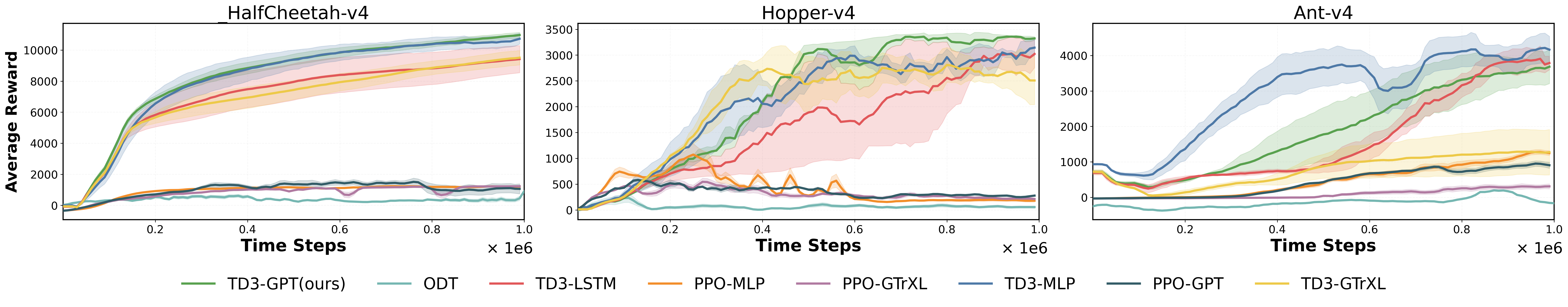}
    \caption{Performance on selected MuJoCo tasks. We train GPT and GTrXL using both PPO and TD3 to ensure fair evaluation.}
    \label{fig:exp_1_mujoco}
\end{figure*}

\section{Comparison with Baselines}

To assess the strength of our transformer setup, we implement all practical takeaways in complete TD3, PPO, and SAC agents and compare them with strong baselines: LSTM, MLP, GTrXL, ODT, and CNN variants. RL parameters are kept fixed across runs (Appendix~A, Tables~\ref{opt_params} and~\ref{tab:params_sac}). 
TD3 and PPO are from CleanRL~\citep{huang2022cleanrl} using default hyperparameters, while SAC follows the original ManiSkill3 benchmark settings.

\subsection{MuJoCo MDP Environments}
For MDPs, we apply the \texttt{\textbf{ObsOnly}} conditioning from RQ1, cross-episode slicing from RQ3, and separate backbones from RQ2 to avoid gradient interference.

Figure~\ref{fig:exp_1_mujoco} shows that TD3-GPT performs on par with TD3-MLP and TD3-LSTM across all tasks, achieving stable and strong learning. This confirms that transformers can effectively process sequential input even in fully observable settings. In contrast, PPO fails to train both GPT and MLP agents, suggesting it is suboptimal in this setting. Transformer-based baselines like GTrXL and ODT also perform poorly. Although ODT is not fully online, our approach outperforms it while relying solely on online data. For fair comparison, we evaluate GTrXL using both TD3 and PPO.

\subsection{MuJoCo POMDP Environments}

\begin{figure}[b!]
    \centering
    \includegraphics[width=1\linewidth]{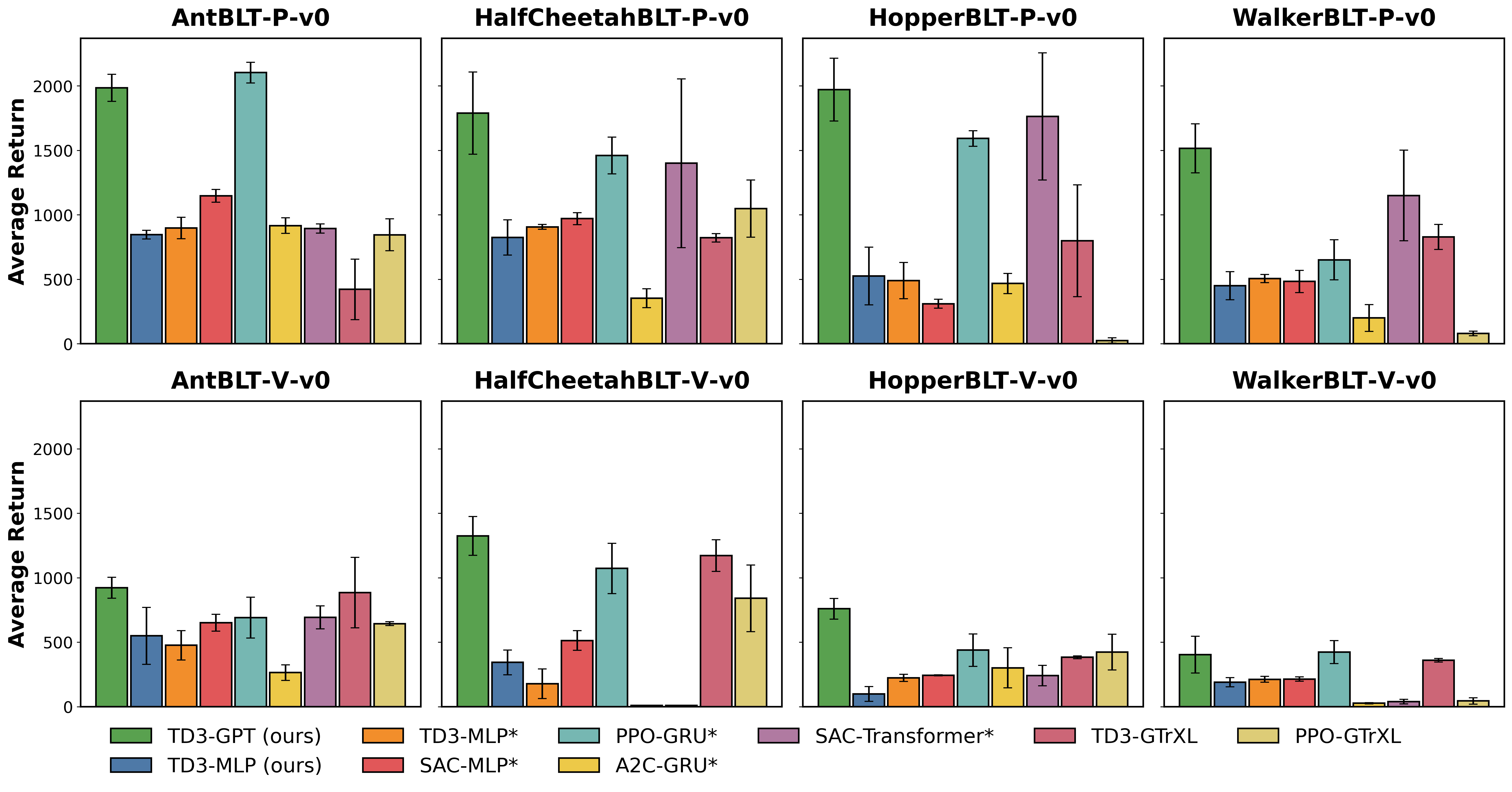}
     \caption{POMDP MuJoCo tasks with masked positions (left) and masked velocities (right). * – results from \citep{luo2024efficientrecurrentoffpolicyrl}.}
     \label{fig:pomdp_bars}
\end{figure}

With partial observability, we switch from \texttt{\textbf{ObsOnly}} to \texttt{\textbf{EmbedConcat}} conditioning, retain separate actor and critic backbones, and use TD3 for training. As shown in Figure~\ref{fig:pomdp_bars}, TD3-GPT outperforms MLPs, RNNs, and vanilla SAC-Transformer baselines. GTrXL is the strongest alternative, but still lags behind our model in most tasks.

\subsection{Vector-based ManiSkill3 Environments}

For the vector-based ManiSkill3 environments, we trained a GPT-based agent with SAC using a context size of 10. Due to the Markovian nature of ManiSkill3 tasks, we use the \texttt{\textbf{ObsOnly}} conditioning approach with cross-episode slicing and first observation masking (see Appendix~C for more experiments with masking). 
Table~\ref{tab:vec_ms} shows that the transformer-based model achieves performance comparable to the MLP baseline and in some cases even surpasses it. In turn, the LSTM baseline, trained under the same setup, achieved similar overall performance, except on the \texttt{PushT} task, where it underperformed. These results suggest that the transformer is a competitive baseline for this class of continuous control tasks.

\begin{table}[H]
  \centering
  \small
  \caption{GPT performance on vector-based ManiSkill3 tasks.}
  \label{tab:vec_ms}
  % \begin{adjustbox}{width=1\columnwidth}
  \begin{tabular}{lccc}
    \toprule
    {} & \texttt{\textbf{PushT}} & \texttt{\textbf{PickCube}} & \texttt{\textbf{TriFingerRotateCube}} \\
    \midrule
    SAC-GPT  & $0.65\pm0.02$ & $0.99\pm0.0$ & $0.94\pm0.0$ \\
    SAC-MLP  & $0.59\pm0.01$ & $0.99\pm0.0$ & $0.85\pm0.0$ \\
    SAC-LSTM & $0.22\pm0.01$ & $0.94\pm0.0$ & $0.90\pm0.0$ \\
    \bottomrule
  \end{tabular}
  % \end{adjustbox}
\end{table}

\subsection{Image-based ManiSkill3 Environments}

An important component of our research is assessing the effectiveness of the proposed takeaways in image-based environments, as real-world manipulation tasks typically rely on camera-equipped robots. In this experiment, we evaluated a transformer-based agent on image-based versions of several ManiSkill environments (\texttt{Push Cube}, \texttt{PickCube}, and \texttt{PokeCube}). Transformer training parameters were kept identical to those used in the vector-based setting, including ObsOnly conditioning with the first observation masking.

To process visual input, we used a lightweight convolutional encoder composed of convolutional layers with ReLU activations, interleaved with max-pooling operations, progressively reducing spatial resolution while increasing channel depth up to 64.
All other transformer training parameters remained identical to those used in the vector-based setting.

\begin{table}[t]
  \centering
  \small
  \caption{GPT performance on image-based ManiSkill tasks.}
  \label{tab:img_ms}
  \begin{tabular}{lccc}
    \toprule
    {} & \texttt{\textbf{PushCube}} & \texttt{\textbf{PickCube}} & \texttt{\textbf{PokeCube}} \\
    \midrule
    SAC-GPT & $0.99\pm0.0$ & $0.97\pm0.02$ & $0.86\pm0.05$ \\
    SAC-CNN & $0.99\pm0.0$ & $0.98\pm0.01$ & $0.61\pm0.02$ \\
    \bottomrule
  \end{tabular}
\end{table}

Table~\ref{tab:img_ms} demonstrates the transformer's ability to learn effectively even in image-based environments, demonstrating that our practical takeaways extend to realistic robotic settings where agents rely on visual observations.

%%%%%%%%%%%%%%%%%%%%%%%%%%%%%%%%%%%%%%%%%%%%%%%%%%%%%%%%%%%%%%%%%%%%%%%%

\section{Conclusion}
In this work, we re-evaluate the use of transformers in online reinforcement learning and position them as strong baselines capable of matching or surpassing MLPs, LSTMs, CNNs, and the GTrXL variant across continuous control tasks. Through extensive experiments guided by targeted research questions, we analyzed the inner workings of transformer-based agents and distilled practical takeaways to inform future applications in online RL. By consolidating these insights into a unified training setup, we demonstrated strong performance and stable learning across diverse environments. This reframes the transformer from a notoriously unstable model into a competitive and easy-to-train alternative to common baselines.

We believe our work offers valuable insights into the training dynamics of transformers in online RL and lays a foundation for future advancements in the field.

%%%%%%%%%%%%%%%%%%%%%%%%%%%%%%%%%%%%%%%%%%%%%%%%%%%%%%%%%%%%%%%%%%%%%%%%
% \clearpage

\appendix

\section{Appendix A -- Training Details}
\label{sec:reference_examples}

\vspace{5ex}  

\begin{table}[htbp]
  \centering
  \caption{Hyperparameters for TD3-based baselines.}
  \label{opt_params}
  \small
  \begin{sc}
    \begin{tabular}{lcc}
      \toprule
      \textbf{Parameter} & \textbf{TD3-GPT} & \textbf{TD3-MLP} \\
      \midrule
      $\gamma$-discount      & 0.99 & 0.99 \\
      $\tau$-soft update     & 0.005 & 0.005 \\
      Policy noise           & 0.2 & 0.2 \\
      Noise clip             & 0.5 & 0.5 \\
      Exploration noise      & 0.1 & 0.1 \\
      Batch size             & 256 & 256 \\
      Learning Rate          & $3\times10^{-4}$ & $3\times10^{-4}$ \\
      Buffer size            & $1.5\times10^{6}$ & $1.5\times10^{6}$ \\
      Learning Starts        & 25000 & 25000 \\
      Seeds                  & 1,2,3,4 & 1,2,3,4 \\
      \midrule
      Num layers             & 1 & 2 \\
      Num heads              & 4 & -- \\
      Dim model              & 128 & -- \\
      Dim feedforward        & 256 & 256 \\
      Dropout                & 0.0 & -- \\
      Context len            & 10 & -- \\
      \bottomrule
    \end{tabular}
  \end{sc}
\end{table}
\begin{table}[!htbp]
  \centering
  \caption{Hyperparameters for SAC-based baselines.}
  \label{tab:params_sac}
  \small
  \begin{sc}
    \begin{tabular}{lccc}
      \toprule
      \textbf{Parameter} & \textbf{SAC-GPT} & \textbf{SAC-LSTM} & \textbf{SAC-MLP} \\
      \midrule
      $\gamma$ (discount)     & 0.8  & 0.8  & 0.8 \\
      $\tau$ (tau)            & 0.01 & 0.01 & 0.01 \\
      $\alpha$ (alpha)        & 0.2  & 0.2  & 0.2 \\
      Update epochs           & 10   & 10   & 10 \\
      Batch size              & 1024 & 1024 & 1024 \\
      Learning starts         & 4000 & 4000 & 4000 \\
      Learning rate           & $3\times10^{-4}$ & $3\times10^{-4}$ & $3\times10^{-4}$ \\
      Seeds                   & 1,2,3,4,5,6 & 1,2,3,4,5,6 & 1,2,3,4,5,6 \\
      \midrule
      Num. layers             & 1   & 1   & 3 \\
      Num. heads              & 2   & --  & -- \\
      Dim. model              & 256 & 256 & 256 \\
      Dim. feedforward        & 512 & 512 & 512 \\
      Context length          & 10  & 10  & -- \\
      \bottomrule
    \end{tabular}
  \end{sc}
\end{table}
\begin{table}[!htbp]
  \centering
  \caption{Hyperparameters for PPO-based baselines.}
  \label{tab:params_all}
  \small
  \begin{sc}
    \begin{tabular}{lcc}
      \toprule
      \textbf{Parameter} & \textbf{PPO-GPT} & \textbf{PPO-MLP} \\
      \midrule
      $\gamma$-discount  & 0.99 & 0.99 \\
      GAE $\lambda$      & 0.95 & 0.95 \\
      Max grad norm      & 0.5  & 0.5 \\
      Clip coef          & 0.2  & 0.2 \\
      Vf coef            & 0.5  & 0.5 \\
      Update epoch       & 10   & 10  \\
      Num minibatches    & 32   & 32  \\
      Batch size         & 2048 & 2048 \\
      Learning Rate      & $3\times10^{-4}$ & $3\times10^{-4}$ \\
      Seeds              & 1,2,3,4 & 1,2,3,4 \\
      \midrule
      Num layers         & 1 & 3 \\
      Num heads          & 4 & -- \\
      Dim model          & 128 & -- \\
      Dim feedforward    & 128 & 64 \\
      Context len        & 10 & -- \\
      \bottomrule
    \end{tabular}
  \end{sc}
\end{table}
\begin{table}[htbp]
  \centering
  \caption{Hyperparameters for the scaling experiment for model sizes: 51k, 320k, 1M, 5M}
  \label{tab:params_scaling}
  \small
  \begin{sc}
    \begin{tabular}{lcc}
      \toprule
      \textbf{Parameter} & \textbf{TD3-GPT} & \textbf{TD3-MLP} \\
      \midrule
      $\gamma$-discount      & 0.99 & 0.99 \\
      $\tau$-soft update     & 0.005 & 0.005 \\
      Policy noise           & 0.2 & 0.2 \\
      Noise clip             & 0.5 & 0.5 \\
      Exploration noise      & 0.1 & 0.1 \\
      Batch size             & 256 & 256 \\
      Learning Rate          & $3\times10^{-4}$ & $3\times10^{-4}$ \\
      Buffer size            & $1.5\times10^{6}$ & $1.5\times10^{6}$ \\
      Learning Starts        & 25000 & 25000 \\
      Seeds                  & 1,2,3,4 & 1,2,3,4 \\
      \midrule
      Num layers             & 1, 1, 1, 6& 2, 5, 4, 5 \\
      Num heads              & 4, 4, 4, 4& -- \\
      Dim model              & 32, 128, 256, 256 & -- \\
      Dim Backbone        & 256 & 256, 512, 1024 \\
      Dropout                & 0.0 & -- \\
      Context len            & 10 & -- \\
      \bottomrule
    \end{tabular}
  \end{sc}
\end{table}

\begin{figure*}[t!]%[t]
    \centering
    \includegraphics[width=1\textwidth]{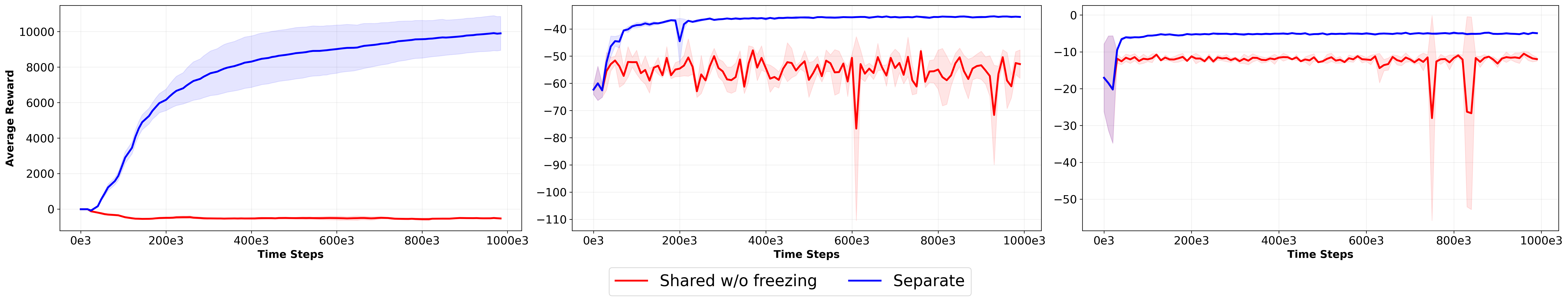} 
    \caption{Comparison of different TD3-MLP architectures on \texttt{HalfCheetah} (left), \texttt{Pusher} (center) and \texttt{Reacher} (right).}
    \label{fig:exp_4}
\end{figure*}

This section presents the parameters used for training the models. For the MuJoCo and MuJoCo-POMDP tasks, we employed PPO and TD3-based implementations; the model parameters are shown in Tables~\ref{opt_params}, \ref{tab:params_all}. For the ManiSkill3 tasks, we used SAC, with the parameters given in Table \ref{tab:params_sac}.

\section{Appendix B -- Additional RQs}

This section presents additional research questions that were not directly raised earlier but may serve as a useful supplement and address questions that could arise when reading the paper.

\subsection{Additional RQ1: How stable are transformers as they scale compared to MLPs?}
\label{sec:scaling}
% \textbf{Research Question.}
% How well do transformers scale compared to MLPs in terms of performance, and should we expect a loss of stability from larger models?

\textbf{Details.}
To evaluate the change in performance of transformers and MLPs as the number of model parameters increases with fixed RL parameters, we tested two variations: TD3-MLP and TD3-GPT. We fixed the RL parameters across both models and increased: the number of transformer or MLP layers and the hidden dimension (see Table~\ref{tab:params_scaling}). As a result, both models had a comparable number of parameters (51k, 320k, 1M, 5M). We then tested these models on two standard MuJoCo environments.

\begin{figure}[h]
    \centering
    \includegraphics[width=1.0\linewidth]{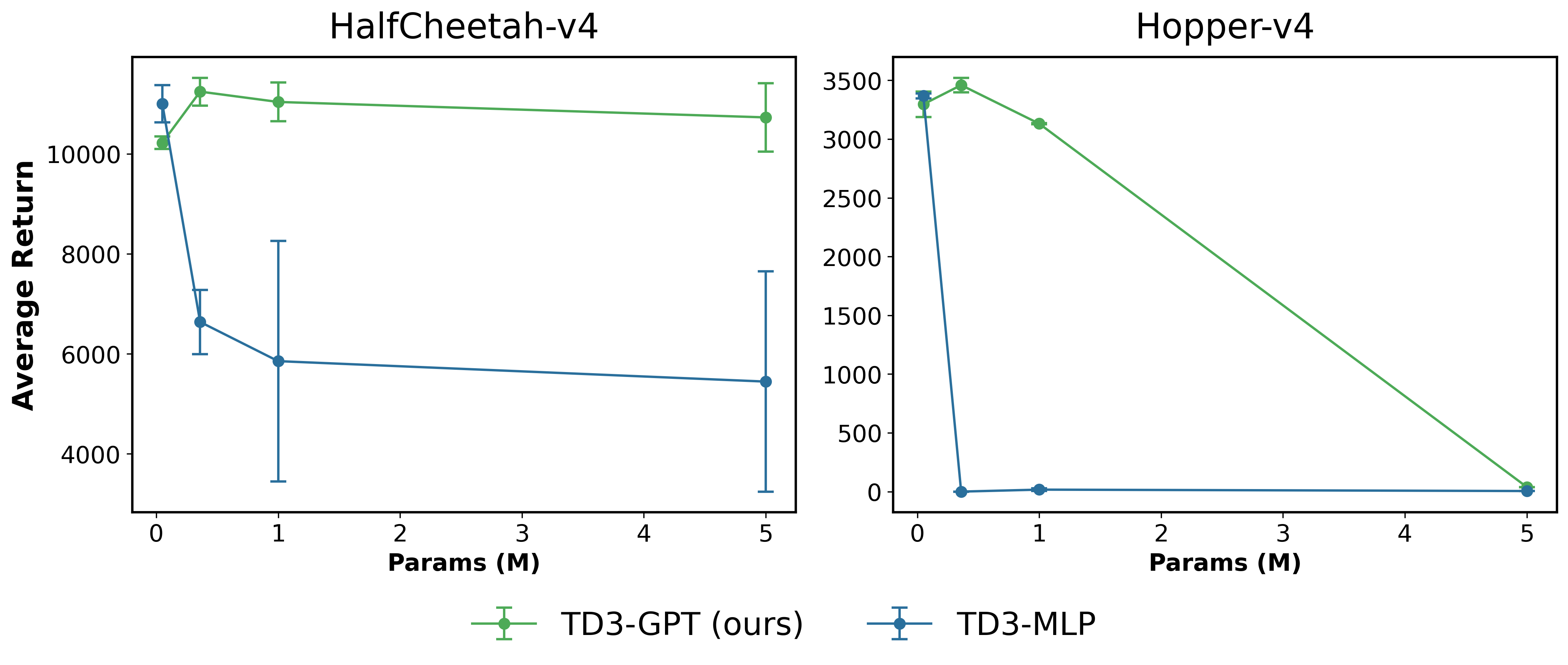} 
    \caption{Comparison of performance of TD3-GPT and TD3-MLP on \texttt{HalfCheetah-v4} and \texttt{Hopper-v4} for different parameters number}
    \label{fig:exp_2}
\end{figure}

\textbf{Interpretation.}
Our results showed that increasing the number of parameters in MLP-based models caused them to fail to learn properly under the original training settings. As seen in Figure~\ref{fig:exp_2}, learning quality significantly degraded. In contrast, transformers were less affected by the parameter increase, showing higher performance. 

\begin{tcolorbox}[colback=white!5!white, colframe=black, width=\linewidth, arc=5pt]
\textbf{Practical Takeaway:} Transformers are more robust to parameter scaling rather than MLPs: MLP-based models does not scale well without careful tuning.
\end{tcolorbox}

%########################################################
%#############  Shared/Separate ENCODER #################
%########################################################

\subsection{Additional RQ2: Is there a sharing issue with MLP backbones?}
\label{sec:share_separate_mlp}
% \textbf{Research Question.}
% Previous research has reported similar issues with sharing in RNN-based off-policy architectures \citep{ni2023transformers}. Is this a general issue across all sequential models, or does it extend beyond them? Do such problems also arise with MLP-based encoders?

\textbf{Details.}
To conduct this experiment, similarly to the previous one but with MLP, we added an additional linear layer to the TD3-MLP architecture for preprocessing observations. We then tested two settings: 
1. Shared MLP Encoder Layer – The actor and critic share an MLP encoder layer, which both can update using their gradients. 2. Separate MLP Encoder Layer – The actor and critic each have their own MLP encoder layer for processing observation sequences.

\textbf{Interpretation.} Results in Figure~\ref{fig:exp_4} show that the issues associated with encoder sharing are not exclusive to sequential models; they also arise in MLP architectures. Our experiment demonstrated that similar learning instabilities occur when an MLP agent shares an encoder without freezing it.

The experiment confirmed that the issues previously observed in RNNs and transformers with shared encoders also manifest themselves in MLPs. When the actor and critic share an MLP encoder and update it with their gradients, conflicts in learning arise, leading to convergence degradation.

\begin{tcolorbox}[colback=white!5!white, colframe=black, width=\linewidth, arc=5pt]
\textbf{Practical Takeaway:} The sharing issue is not unique to sequential models (RNNs, Transformers) but is also present in MLPs. This indicates a fundamental difficulty in jointly training the actor and critic in off-policy algorithms with the same feature extraction function in state based environments.
\end{tcolorbox}

\clearpage

\end{document}